\documentclass[sigconf]{acmart}

\usepackage{balance}

\AtBeginDocument{%
  \providecommand\BibTeX{{%
    \normalfont B\kern-0.5em{\scshape i\kern-0.25em b}\kern-0.8em\TeX}}}

\begin{document}

\fancyhead{}

\title{I Know What You Would Like to Drink: \\ Benefits and Detriments of Sharing Personal Info\\ with a Bartender Robot}


\author{Alessandra Rossi*,\hspace{1em}
Vito Giura**, \hspace{1em}
Carmine Di Leva**, \hspace{1em}
Silvia Rossi*\hspace{1em}}

\affiliation{%
 \vspace{0.5em}\institution{University of Naples Federico II}
  \city{Naples}
  \country{Italy}\\
  \institution{*\{alessandra.rossi,\hspace{0.2em}silvia.rossi\}@unina.it,\hspace{0.2em} **\{vi.giura, car.dileva\}@studenti.unina.it}
}


\begin{abstract}
This paper introduces benefits and detriments of a robot bartender that is capable of adapting the interaction with human users according to their preferences in drinks, music, and hobbies. We believe that a personalised experience during a human-robot interaction increases the human user's engagement with the robot and that such information will be used by the robot during the interaction. However, this implies that the users need to share several personal information with the robot. In this paper, we introduce the research topic and our approach to evaluate people's perceptions and consideration of their privacy with a robot. We present a within-subject study in which participants interacted twice with a robot that firstly had not any previous info about the users, and, then, having a knowledge of their preferences. We observed that less than 60\% of the participants were not concerned about sharing personal information with the robot.
\end{abstract}


\maketitle

\section{Introduction}
\label{sec:intro}
Nowadays, service robots have been employed in different public environments, such as restaurants, fairs, museums, and hospitals. An application context for service robots that is particularly challenging is the bartending domain \cite{10.5555/3038718.3038772}. A robot deployed in human populated environments needs not only to be able to successfully complete a task, but also to show socially intelligence to enhance its interaction with a robot \cite{Salem2013}. Bartender robots need to manipulate objects, and actively engage customers by personalising the interactions in order to make them more pleasant and long-lasting. 

Such robots need to acquire a wide range of information regarding their users (e.g. personal details, preferred drinks, hobbies, habits) in order to adapt their behaviours to the human users' needs and preferences. This information can be used to suggest preferred or new drinks and personalise the interaction with music and topics to be verbally introduced while preparing them. However, while privacy issues can be addressed by encrypting the data collected or when possible reducing the amount of data stored \cite{Basharat2012DatabaseSA,10.1145/1749603.1749605}, users remain very concerned of sharing such data \cite{10.1145/3171221.3171269,Syrdal2007}. Moreover, since the data could be used to personalise the interaction in a public space, this could produce a breach of confidentiality. These might consequently affect negatively users' interaction with a robot. 
In this work, we present an exploratory study to investigate whether people's perception of privacy affects their willingness to sharing personal information with a robot bartender.

\section{Approach}
\label{sec:method}



The focus of this study was to evaluate the impact of a personalised interaction with a robot to build a long-term relationship between people and a service robot. In particular, we were interested in the people's perception of privacy when sharing personal information with a robot.

The study was organised as a within-subject experimental design. 
We conducted an online video study where each participant watched two different interactions with the robot. The first time, the robot bases its interaction with the participants without having any previous information of the user. During the second interaction, the robot personalises the interaction according to the information collected during the first interaction.

In the video, an actor used a magnetic card to register and, later, to be recognised by the robot, and, finally, simulated the payment for a drink. The actor interacted with a robotic arm using voice commands. 

\subsection{Procedure}

Actors were given a magnetic card and asked to swipe it on a RFID reader. During their first interaction with the robot, the actors were able to register their personal information (i.e. name, age, gender), and preferences (i.e. drinks, hobbies and music genre) using a computer to be used for personalising the second interaction. Then, they were able to order a drink from the robot by speech using an external microphone connected through Bluetooth to the robot. The participant's chosen drink is also stored in their magnetic card to be used during the second interaction. 

During the second interaction, actors were recognised swiping their magnetic card, and the robot suggested preparing the last ordered drink for them while playing a song from their favourite music genre. The robot also engaged participants by talking with them about their favourite hobbies.

In both interactions, the graphical application on the computer was designed to improve the interaction using colours that convey serenity and calm to the user \cite{Connor2011}. We also decided to use a simple robotic arm with a synthetic voice to be able to meet user expectations. While people are becoming very familiar with virtual assistant AI technologies, such as Amazon Alexa\footnote{Amazon Alexa \url{https://www.amazon.com}}, their expectation of robot capabilities have also raised. However, the absence of the robot's ability to sense and respond to the user's actions and intentions had been a limiting factor in the success of an HRI, it specifically affects people's trust, in the robot \cite{9223471}. For this reason, we decided to use a simple robotic arm that is able to speak to the user and prepare a drink.

At the end of each interaction, we asked participants to complete a short questionnaire.

\subsection{Measures}
Participants were asked whether they had any previous experience of interaction with robots. Moreover, we asked participants to confirm that they were truly involved in the study by asking them which was the cocktail suggested by the robot, and finally ordered by the actor. 

At the end of each video, participants were asked to evaluate the usability, the level of sympathy, security and intelligence of the robot. In particular, we asked the participants to rate their perception of privacy by providing personal information to the robot.

\subsection{Participants}
We recruited 78 participants (45 male, 33 female), aged between 17 and 59 years (avg. 25.7\%, dev. std. 6.90\%). Three participants failed the engagement test, i.e. those who gave more than one wrong answer thus identified as not paying very much attention to the study, which can be expected in an online survey \cite{Berinsky2011}. We, therefore, decided to exclude them from further analysis.

\section{Results}
\label{sec:res}


The majority of participants (74.4\%) declared to have never had any experience with a robot, while the remaining had some interactions with robots. These participants also stated to have medium-high expectations of the robot's capabilities, expectations that were also confirmed for the 56\% of participants, and increased for the 40\% of participants after the second interaction.

We observed that 50.7\% participants would have preferred to be served by a robot with more human-like appearances, while only the 20\% was content of the proposed robot's appearance. This might be due to the fact that a robot with human-like physical features might be perceived to be more intelligent than one without \cite{10045_12585,Rossi2018b}. 

57.3\% of participants rated to be comfortable providing sensible information to the robot, the 25.3\% were uncomfortable to share the personal information, and the remaining were neither comfortable nor uncomfortable. 

Finally, we observed that participants felt safe to interact with the robot with average ratings 3.77 over a scale from 1 (very unsafe) to 5 (very safe).

\section{Future works / Conclusions}
\label{sec:conc}
Our main research interest is focused on investigating how to deploy a service robot that is able to engage users in a dynamic and unsupervised interaction. In particular, we are interested in investigating how a long-term interactive relationship can be established and preserved between human users and a service robot personalising people's experience and adapting the robot's behaviours to people's needs and requests. The results presented in this study will serve as preliminary study to further investigate people's willingness to share personal information with a robot to have a personalised experience with a robot. We will also further investigate how the robot's appearance affects people's perception of the robot and their willingness to share information.

\section{Acknowledgments}
This work has been supported by Italian PON I\&C 2014-2020 within the BRILLO research project “Bartending Robot for Interactive Long Lasting Operations”, no. F/190066/01-02/X44.

\bibliographystyle{ACM-Reference-Format}
\balance 
\bibliography{biblio}

\end{document}